\documentclass[times,review,10pt]{elsarticle}
\biboptions{numbers,sort&compress}

\usepackage{iftex}
\usepackage{amsmath,amssymb}
\usepackage{geometry}
\usepackage{setspace}
\usepackage{graphicx}
\usepackage{url}
\usepackage{bm}
\usepackage{xcolor}
\usepackage[colorlinks]{hyperref}
\usepackage{array}
\usepackage{booktabs}
\usepackage{multirow}
\usepackage{placeins}
\usepackage{xspace}
\usepackage{ragged2e}
\usepackage[font={sf,footnotesize},labelfont=bf]{caption}
\ifPDFTeX
  \usepackage[T1]{fontenc}
  \usepackage[utf8]{inputenc}
  \usepackage{textcomp}
\else
  \usepackage{fontspec}
\fi
\AddToHook{env/table/begin}{\normalsize}
\AddToHook{env/table*/begin}{\normalsize}
\makeatletter
\AddToHook{cmd/@footnotetext/before}{\fontsize{8pt}{10pt}\selectfont}
\AddToHook{cmd/@mpfootnotetext/before}{\fontsize{8pt}{10pt}\selectfont}
\makeatother

\makeatletter
\renewcommand{\fnum@figure}{Fig.~\thefigure}
\makeatother

\newcommand{\etal}{\textit{et al.}}

\newcommand{\PaperFullTitle}{EAPFusion: Intrinsic Evolving Auxiliary Prior Guidance for Infrared and Visible Image Fusion}
\newcommand{\MethodName}{EAPFusion\xspace}

\begin{document}
\justifying
\pagestyle{plain}
\let\WriteBookmarks\relax
\renewcommand{\floatpagefraction}{0.8}
\renewcommand{\textfraction}{0.1}
\renewcommand{\topfraction}{0.9}
\renewcommand{\bottomfraction}{0.8}
\setcounter{topnumber}{2}
\setcounter{bottomnumber}{1}
\setcounter{totalnumber}{3}

\begin{frontmatter}
\title{\PaperFullTitle}

\author[a]{Zhenyu Sun}
\ead{sunzhenyu@mail.nwpu.edu.cn}
\author[a]{Luobin Zhang}
\ead{zhangluobin@mail.nwpu.edu.cn}
\author[a]{Axi Niu}
\ead{nax@nwpu.edu.cn}
\author[a]{Haishen Wang}
\ead{wanghaishen@mail.nwpu.edu.cn}
\author[a]{Qingsen Yan\corref{cor1}}
\ead{qingsenyan@nwpu.edu.cn}
\cortext[cor1]{Corresponding author}
\address[a]{Northwestern Polytechnical University, Xi'an, China}

\begin{abstract}
\sloppy
Infrared-visible image fusion aims to create an information-rich fused image by integrating the complementary thermal saliency from infrared sensing and fine textures from visible imaging. Such accurate fusion is essential for real-world perception applications in complex scenes, including nighttime autonomous driving, search and rescue, and surveillance, and can further benefit downstream tasks such as semantic segmentation. However, most existing fusion methods rely on static trained weights that cannot adapt to scene-specific content at inference time, and often suffer from a granularity mismatch when coarse auxiliary semantics are injected, which makes it difficult to simultaneously highlight targets and preserve details. In this work, we propose \MethodName to address these issues by using self-evolving intrinsic priors instead of relying on external auxiliary models. Concretely, \MethodName maintains a compact set of intrinsic priors and progressively updates them across scales. These evolved priors are utilized to dynamically generate convolutional kernels, shifting the paradigm from fixed, pre-trained filters to instance-adaptive parameters via prior-conditioned dynamic convolution. Furthermore, we design a channel-level fusion module that shuffles and interleaves infrared and visible channels, applying local channel mixing to boost cross-modal complementarity. Experiments on different datasets, including cross-dataset evaluation and semantic segmentation, show that the proposed method achieves state-of-the-art quantitative and qualitative fusion results, and consistently boosts downstream performance. Code is coming soon.
\fussy
\end{abstract}

\end{frontmatter}

\section{Introduction}
\label{sec:introduction}
Complex real-world scenarios often involve multiple factors, such as low illumination, occlusion, smoke and haze, and strong background interference, making it difficult for a single imaging sensor to simultaneously ensure target visibility, texture details, and imaging robustness~\cite{Liu_2024_DSfusion}. Accordingly, multimodal image fusion has emerged as a key technique for all-weather perception: it integrates information from complementary sensors into a single, more informative and interpretable image representation, benefiting both human observation and downstream machine vision tasks. Among various fusion settings, infrared and visible image fusion (VIF) is particularly representative. Infrared imaging reflects the spatial distribution of thermal radiation and can highlight salient targets at night or under adverse weather, but it is typically limited by low spatial resolution and poor textural fidelity, leading to insufficient details and contrast. Conversely, visible images contain richer spatial details and structural textures, yet are more sensitive to illumination changes and occlusion. Therefore, VIF aims to generate fused results that combine infrared target saliency with clearly preserved visible details, and has shown significant value in applications such as nighttime autonomous driving, search and rescue, and security surveillance~\cite{Luo_2024_fullscale}.

Recent deep learning approaches for VIF generally fall into two paradigms. The first line targets image reconstruction and low-level fidelity, typically employing CNN-based autoencoders~\cite{Zhao_2020} for local texture extraction or Transformers~\cite{Liu_2026_sourceInteraction} for long-range dependency modeling. Nevertheless, since their optimization objectives primarily emphasize signal reconstruction and statistical preservation, these methods inherently struggle to capture high-level semantic information. To compensate for this semantic deficiency, an auxiliary-guided fusion paradigm has emerged~\cite{Zheng_2025_detectionGuided}. Specifically, these methods incorporate external teacher models, leveraging pre-trained networks for high-level tasks like object detection~\cite{Guo_2025_SpTFuse}, or utilizing CLIP to transform natural language into steering features~\cite{Zhang_2025_CLIPBasedLowRedundancy}. While effectively endowing fusion networks with semantic cues, the heavy reliance on these external priors introduces two inherent tensions that constrain practical fusion performance.

First, regarding the feature interaction mechanism, most existing networks rely on fixed parameters, i.e., a static processing scheme independent of the input content. During inference, such mechanisms cannot adaptively adjust convolutional kernel weights according to the characteristics of the current sample (e.g., illumination intensity, texture complexity, or target saliency). Using a single fixed interaction pattern for diverse scenes makes it difficult for the model to adapt to sample-specific properties, resulting in either suppressed IR targets or the loss of VIS details.

Second, there exists a granularity mismatch between auxiliary semantics and low-level texture requirements. Features extracted by external auxiliary models are typically high-level semantics that focus on abstract object-category information, whereas the shallow layers of a fusion network mainly extract low-level textures and edge details~\cite{Zhao_2024_ICML}. Many existing methods ignore this discrepancy and directly inject high-level semantics into shallow layers. Introducing coarse-grained semantics into fine-grained feature layers may interfere with low-level texture representations, causing high-frequency detail loss or visual artifacts.

To address the above issues, we propose \MethodName. The core idea is to construct an internal guidance mechanism that eliminates the dependence on external auxiliary models, enabling the network to automatically mine high-value auxiliary priors from its own features.

Specifically, to mitigate the mismatch between high-level semantics and low-level details, we propose an Adaptive Prior Generator (APG) which establishes an auxiliary-prior updating mechanism evolving synchronously with the backbone. Unlike the static injection of external semantics, APG adopts a progressive absorption strategy: it inherits historical priors while continuously updating them with current-level features, allowing the auxiliary information to transition smoothly from shallow to deep layers and thereby alleviating conflicts caused by cross-level granularity inconsistency.

Furthermore, to overcome the static nature of interaction mechanisms at inference time, we introduce the Prior-Driven Dynamic Convolution Block (DDCB). This block transforms the hierarchical evolution of priors into test-time adaptive interactions: it generates dynamic interaction rules according to the input content (e.g., illumination or texture structures), shifting feature aggregation from fixed-parameter fitting to dynamic instance reasoning, and enabling adaptive fusion under varying scenes and imaging conditions.

To better match the front-end encoder, we further design the Shuffle Channel Fusion Block. It shuffles and interleaves IR and VIS channels to facilitate thorough cross-modal interaction along the channel dimension, and then performs effective information integration via local channel mixing.

The main contributions of this paper are summarized as follows:
\begin{itemize}
  \item We propose a self-evolving prior-guided fusion framework that constructs an internal guidance mechanism to autonomously mine high-value auxiliary priors from its own features, eliminating the reliance on external priors.
  \item The proposed APG progressively evolves auxiliary priors from textures to semantics, bridging the granularity gap; meanwhile, the DDCB converts the evolved priors into instance-adaptive interaction rules, shifting feature aggregation from static fitting to dynamic reasoning.
  \item The SCFB leverages channel shuffling and local mixing to facilitate thorough cross-modal interaction along the channel dimension.
  \item Extensive experiments on multiple public datasets demonstrate that our method achieves superior fusion quality and significantly benefits downstream semantic segmentation tasks.
\end{itemize}

\section{Related work}
\label{sec:related-work}
\begin{figure}[!t]
  \centering
  \includegraphics[width=0.98\textwidth]{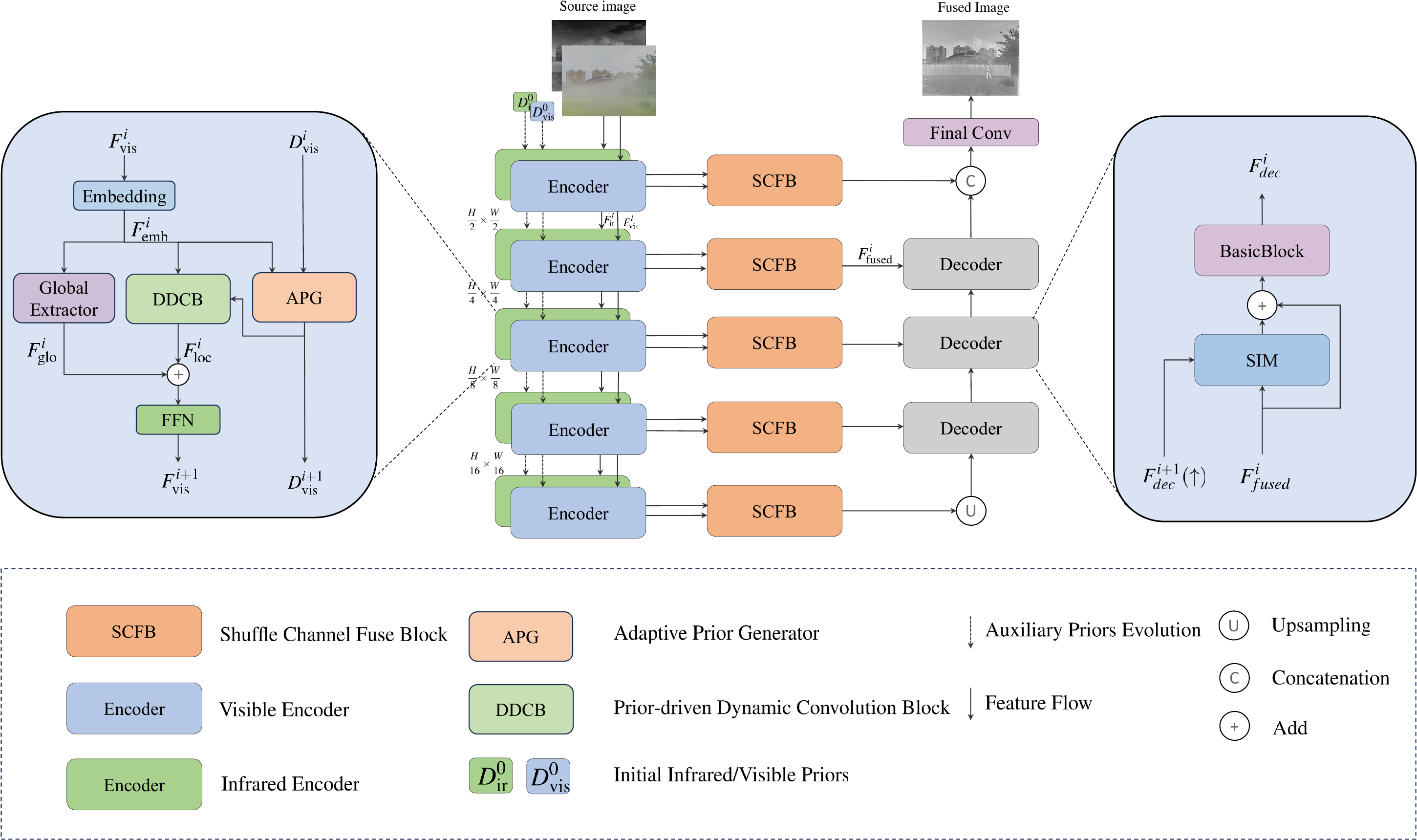}
  \caption{Overall architecture of \MethodName. \MethodName consists of a dual-branch encoder, APG, DDCB, SCFB, and a progressive decoder for infrared-visible fusion.}
  \label{fig:overview}
\end{figure}

\subsection{End-to-End fusion methods}
End-to-end deep networks for infrared--visible image fusion have undergone a systematic architectural evolution, advancing from CNN-based models and generative adversarial networks (GANs) to Transformer architectures and, most recently, diffusion models. CNN-based formulations primarily pursue cross-modal feature aggregation under reconstruction-oriented constraints: Li \etal~\cite{Li_2019} exploited dense feature encoding for intermediate fusion, whereas Li \etal~\cite{Li_2021} devised a residual fusion nesting strategy to preserve fine textures and salient targets. To further sharpen perceptual details, adversarial objectives have been incorporated as implicit realism regularizers, as exemplified by FusionGAN proposed by Ma \etal~\cite{Ma_2019} and the attention-enhanced adversarial design of Li \etal~\cite{Li_2021_AttentionFGAN}. Nevertheless, the convolutional inductive bias inherently restricts long-range dependency modeling, motivating Transformer-based fusion that explicitly parameterizes interactions within and across modalities via attention; Tang \etal~\cite{Tang_2024_ITFuse} proposed an interactive Transformer to facilitate cross-modal information flow, and Ma \etal~\cite{Ma_2022} explored Swin Transformer long-range learning under a unified fusion framework. More recently, diffusion-driven approaches have reframed the fusion process as conditional generation with iterative denoising, where Zhao \etal~\cite{Zhao_2023_DDFM} introduced a denoising-diffusion formulation and Xu \etal~\cite{Xu_2024} extended it with multi-attribute conditioning for improved fidelity.

\subsection{External auxiliary guided methods}
To explicitly align fusion with high-level semantics, external-auxiliary-guided approaches inject task supervision or distill priors from pretrained foundation models. In task-driven settings, fusion is co-optimized with downstream objectives so that the fused output becomes directly discriminative: Liu \etal~\cite{Liu_2022} integrated dual adversarial learning with object-detection training, Bai \etal~\cite{Bai_2025} parameterized learnable fusion losses to emphasize task-critical cues, and Liu \etal~\cite{Liu_2025_DCEvo} formulated fusion--task coupling as an evolutionary multi-objective optimization. Complementarily, language-conditioned fusion has emerged to offer user-controllable semantics: Li \etal~\cite{Li_2024_Text2Pixels} coupled text supervision with detection-oriented training, while Li \etal~\cite{Li_2025} and Liu \etal~\cite{Liu_2025_PromptFusion} leveraged frozen vision--language encoders such as CLIP to align textual concepts with cross-modal visual representations, and Yi \etal~\cite{Yi_2024} introduced semantic text guidance for degradation-aware and interactive fusion. Beyond language, segmentation-centric priors have been distilled from large frozen segmentation models; Wu \etal~\cite{Wu_2025} exploited the Segment Anything Model to provide semantic persistence for multi-modality fusion.

\subsection{Intrinsic prototype priors guided methods}
Intrinsic-prior-based methods attempt to internalize reusable knowledge by maintaining compact prototypes---including memory units, learnable look-up tables, and codebooks---that are repeatedly retrieved and updated to stabilize fusion across scenes. Cheng \etal~\cite{Cheng_2023} proposed MUFusion with a read-write memory and a self-evolution training scheme, while He \etal~\cite{He_2026} introduced a gated recurrent memory mechanism to preserve informative features and suppress redundancy. For efficiency and cross-modality alignment, Yi \etal~\cite{yi2025LUT-Fuse} distilled fusion into learnable look-up tables, Li \etal~\cite{Li_2025_MulFSCAP} explored modality dictionaries for unregistered alignment, and Huang \etal~\cite{Huang_2026_VQDualBranchFusion} investigated vector-quantized codebook representations for robustness and anomaly suppression.

\subsection{Summary}
In summary, existing paradigms face an inherent trade-off between semantic awareness and instance adaptivity: externally guided methods suffer from granularity mismatches, while static prototype models lack content-adaptive expressiveness. To reconcile this tension, we propose \MethodName. Instead of relying on external guidance, \MethodName autonomously mines and progressively evolves internal priors from hierarchical features into dynamic interactions, achieving robust fusion across diverse scenes.

\section{Proposed method}
\label{sec:methodology}

\subsection{Overview}
As shown in Fig.~\ref{fig:overview}, our \MethodName is a self-evolving prior-guided framework for infrared (IR) and visible (VIS) image fusion. It consists of three core components: the Adaptive Prior Generator (APG), the Prior-Driven Dynamic Convolution Block (DDCB), and the Shuffle Channel Fusion Block (SCFB).

\MethodName adopts a dual-branch hierarchical encoder to extract multi-scale representations for the two modalities. At the $i$-th scale, the encoder outputs are denoted as $F_{ir}^{i}$ and $F_{vis}^{i}$, and the network maintains scale-evolving auxiliary priors $D_{ir}^{i}$ and $D_{vis}^{i}$ initialized as $D_{ir}^{0}$ and $D_{vis}^{0}$. Within each scale, an Embedding layer produces embedded features $F_{emb}^{i}$, APG updates auxiliary priors by inheriting history and incorporating current contextual information, and the DDCB then converts these evolved priors into spatially-variant dynamic filters to capture content-adaptive local details. In parallel, a Global Extractor captures global context; the local (DDCB-enhanced) and global representations are aggregated by residual addition and an FFN to produce the next-scale features. SCFB then fuses the two modalities to obtain the scale-wise fused feature $F_{\mathrm{fused}}^{i}$.

The decoder performs progressive reconstruction, as depicted in the right branch of Fig.~\ref{fig:overview}. Starting from the deepest fused feature, each stage performs an upsampling operation, integrates it with the fused feature at the corresponding scale via a semantic injection module (SIM)~\cite{Tang_2023}, and refines the aggregated features with a Basic Block~\cite{Lou_2025}. At the highest resolution, the decoder concatenates shallow details and reconstructs the final fused image $I_f$ with Final Conv. Final Conv is a reconstruction head: it first uses a Basic Block to refine and denoise the high-resolution features and then applies several $3\times3$ convolutions followed by a sigmoid for intensity normalization.

\subsection{Adaptive prior generator }
\begin{figure}[htb]
  \centering
  \includegraphics[width=\linewidth]{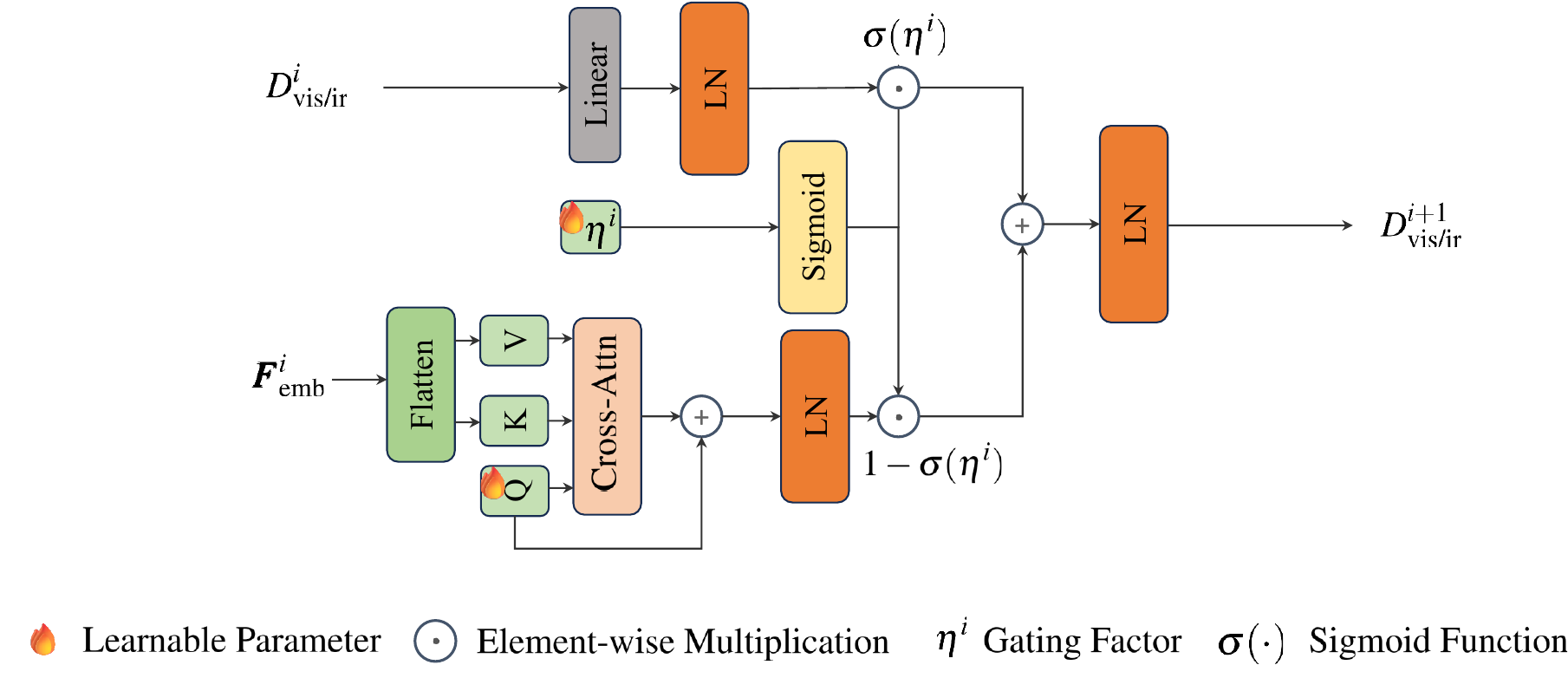}
  \caption{Adaptive Prior Generator (APG). Historical priors are aligned, summarized with current features via cross-attention, and updated by a gated evolution rule.}
  \label{fig:apg}
\end{figure}

In a multi-scale encoder, shallow features tend to capture textures and edges, whereas deeper features are more related to structures and target semantics. If auxiliary priors are injected as static external priors or via a single-shot generation, cross-scale granularity mismatch can easily occur: coarse semantics may disturb shallow layers and harm fine details. To address this issue, we propose an Adaptive Prior Generator (APG), which maintains a compact set of auxiliary priors at each scale and evolves them progressively by inheriting history and incorporating current contextual information. In this way, the auxiliary priors grow in sync with the backbone representations, providing stable and content-related conditioning for subsequent DDCB. As shown in Fig.~\ref{fig:apg}, APG aligns historical priors, summarizes current features via cross-attention, and updates the priors with a gated evolution mechanism.

Given the embedded feature $F_{emb}^{i}$ on scale $i$ and the auxiliary priors from the previous scale $D^{i-1}$, the general APG update is formulated as follows:
\begin{equation}
D^{i}=\mathrm{APG}\big(F_{emb}^{i},D^{i-1}\big),
\label{eq:apg_update}
\end{equation}
where $F_{emb}^{i}\in\mathbb{R}^{B\times C_i\times H_i\times W_i}$ denotes the embedded feature map at the $i$-th scale, $D^{i}\in\mathbb{R}^{B\times M\times C_i}$ denotes the auxiliary priors at this scale, and $M$ is the number of auxiliary priors.

\subsubsection{Initialization and cross-scale inheritance}
We introduce learnable initial priors at the shallowest scale:
\begin{equation}
D^{0}=D_{0},\quad D_{0}\in\mathbb{R}^{M\times C_{1}}.
\label{eq:apg_init}
\end{equation}
We parameterize $D_{0}$ as learnable since the priors will be repeatedly read and updated in subsequent layers. A fixed learnable starting point avoids distribution fluctuations caused by random noise and allows the model to accumulate a set of cross-sample prototype bases during training, improving convergence stability and transferability.

To accommodate the varying channel dimensions across different scales, we align the historical auxiliary priors to the current dimension and normalize them before propagation:
\begin{equation}
\tilde{D}^{i-1}=\mathrm{LN}\Big(\mathcal{A}^{i}\big(D^{i-1}\big)\Big),\qquad
\mathcal{A}^{i}:\mathbb{R}^{M\times C_{i-1}}\rightarrow \mathbb{R}^{M\times C_i},
\label{eq:apg_align}
\end{equation}
\noindent where $\tilde{D}^{i-1}$ represents the aligned and normalized historical auxiliary prior, $\mathcal{A}^{i}(\cdot)$ is a scale-wise dimension alignment mapping, and $\mathrm{LN}(\cdot)$ denotes Layer Normalization

\subsubsection{Extracting candidate priors from current features}
Relying solely on historical inheritance may weaken adaptability to the current content. Therefore, APG also extracts candidate auxiliary features from the current scale features. We first flatten the feature map into a token sequence:
\begin{equation}
K^{i}=\mathrm{Flatten}\big(F_{emb}^{i}\big)\in\mathbb{R}^{B\times N_i\times C_i},\qquad N_i=H_iW_i,
\label{eq:apg_flatten}
\end{equation}
where $\mathrm{Flatten}(\cdot)$ flattens the spatial dimensions $(H_i,W_i)$ into a token sequence while keeping the channel dimension $C_i$, and $N_i=H_iW_i$ is the number of spatial tokens at scale $i$.
Using $M$ learnable query tokens $Q^{i}\in\mathbb{R}^{B\times M\times C_i}$, we perform cross-attention over $K^i$ (with $V^i=K^i$) to produce candidate auxiliary features:
\begin{subequations}\label{eq:apg_ca}
\begin{align}
\Delta D^{i} &= \mathrm{CA}\big(Q^{i},K^{i},V^{i}\big), \label{eq:apg_ca1}\\
D_{\mathrm{prop}}^{i} &= \mathrm{LN}\big(Q^{i}+\Delta D^{i}\big), \label{eq:apg_ca2}
\end{align}
\end{subequations}
where $Q^{i}$ denotes the learnable query tokens at scale $i$, $\mathrm{CA}(\cdot)$ denotes the cross-attention operator, $\Delta D^{i}$ is the residual update extracted from the current features, and $D_{\mathrm{prop}}^{i}$ represents the generated candidate auxiliary prior.

Utilizing $M$ fixed queries to compress information from $N_i$ spatial locations can be viewed as a capacity-controlled summary of key structural patterns at this scale, which is more suitable as conditioning for subsequent dynamic convolution.

\subsubsection{Gated evolutionary update}
Directly replacing the historical priors with $D_{\mathrm{prop}}^{i}$ might discard valuable information from shallow layers. Instead, we adopt a gated, progressive update rule that adaptively balances historical inheritance and the incorporation of current context. Specifically, we formulate this progression as a gated residual update. A learnable gating parameter $\eta^{i}$ is passed through a sigmoid function to dynamically scale the residual between the candidate term $D_{\mathrm{prop}}^{i}$ and the aligned historical term $\tilde{D}^{i-1}$:
\begin{equation}
D^{i} = \mathrm{LN}\Big(\tilde{D}^{i-1} + \sigma(\eta^{i}) \odot \big(D_{\mathrm{prop}}^{i} - \tilde{D}^{i-1}\big)\Big),
\label{eq:apg_gate}
\end{equation}
\noindent where $\sigma(\cdot)$ is the sigmoid function, and $\odot$ denotes element-wise multiplication. A smaller gate value $\sigma(\eta^{i})$ preserves more historical consistency, whereas a larger value integrates more current content for enhanced adaptability.

\subsection{Prior-driven dynamic convolution block}
\begin{figure}[htb]
  \centering
  \includegraphics[width=\linewidth]{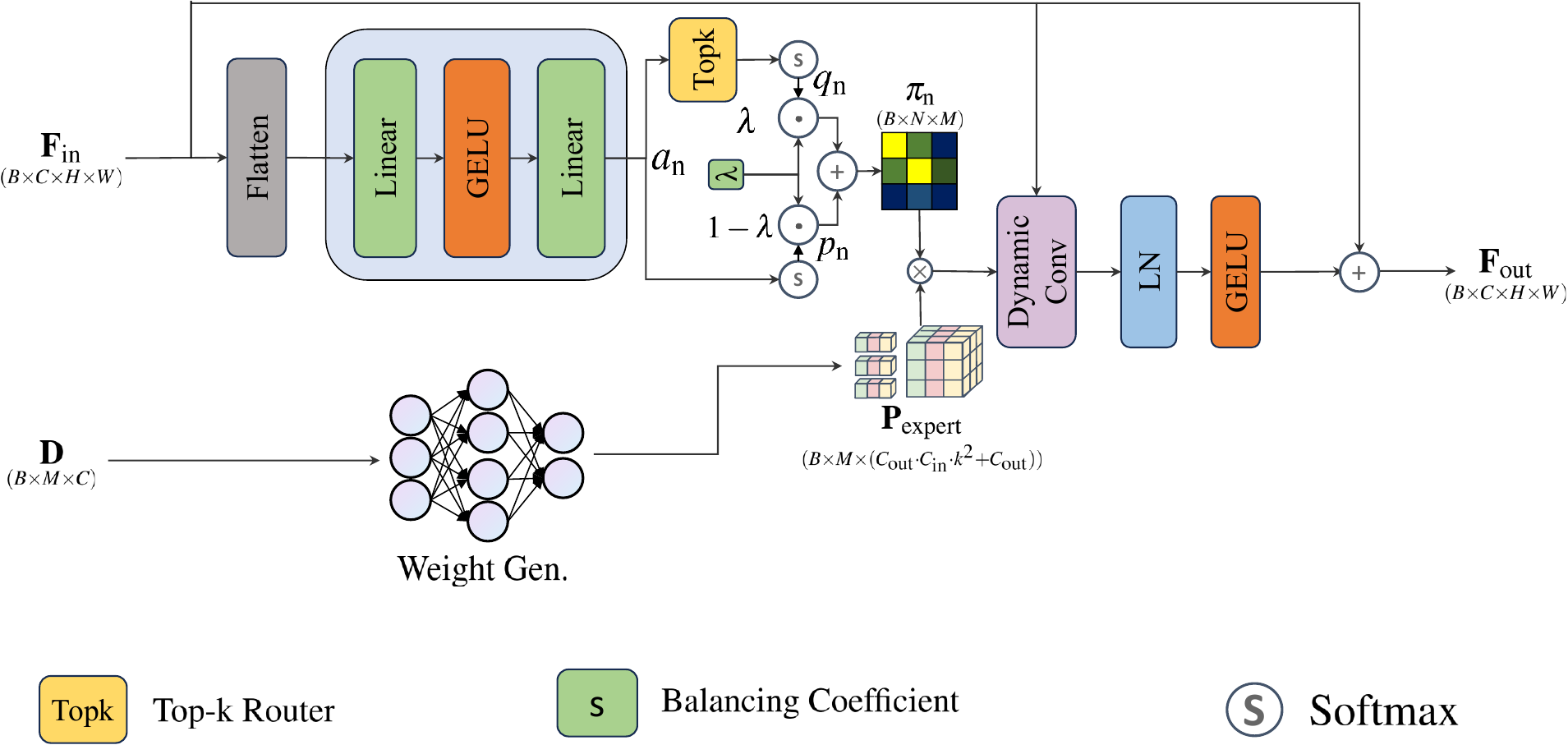}
  \caption{Prior-Driven Dynamic Convolution Block (DDCB). Each prior token generates an expert kernel via WeightGen, and dense/Top-$K$ routing yields location-dependent mixing weights for dynamic convolution.}
  \label{fig:ddc}
\end{figure}

The set of auxiliary priors $D^{i}$ produced by the APG at each scale encapsulates key structural information and salient patterns. However, if they are merely concatenated or injected via simple feature addition, they often fail to induce truly content-adaptive interactions during inference. To overcome this limitation, we transform each auxiliary prior into a unique, content-specific convolution kernel. As illustrated in Fig.~\ref{fig:ddc}, the DDCB achieves this by employing a weight generation network (WeightGen) to parameterize each auxiliary prior into a distinct convolution expert. Subsequently, a hybrid dense and Top-$K$ routing mechanism produces spatially-variant mixing weights. The aggregated expert kernels then execute position-adaptive dynamic convolutions within a residual block. This design facilitates spatially-variant feature extraction, enabling the differentiated enhancement of salient targets and background details.

\subsubsection{Expert generation and routing}
Given an input feature $F_{in}\in\mathbb{R}^{B\times C\times H\times W}$ and a set of priors $D\in\mathbb{R}^{B\times M\times C}$, we map each token $d_m$ to specific convolution parameters via a shared weight generation network (WeightGen):
\begin{equation}
(W_m, b_m) = \mathrm{WeightGen}(d_m), \qquad m=1,\dots,M,
\label{eq:ddc_weightgen}
\end{equation}
\noindent where $W_m$ and $b_m$ represent the generated convolutional kernel and bias parameters for the $m$-th expert, respectively. $d_m\in\mathbb{R}^{C}$ denotes the $m$-th token in $D$, and $M$ represents the total number of experts. This step transforms the descriptive priors into a distinct set of dynamic filter parameters.

At each spatial location $n$, we compute the routing logits with a temperature scaling factor $\tau$:
\begin{equation}
a_{n} = r(f_n) / \tau,
\label{eq:ddc_logits}
\end{equation}
\noindent where $f_n\in\mathbb{R}^{C}$ is the local channel vector at location $n$, $r(\cdot)$ is a lightweight routing network that outputs an $M$-dimensional logit vector, and $a_{n} \in \mathbb{R}^{M}$ represents the unnormalized preference vector over all experts at location $n$. The temperature coefficient $\tau > 0$ controls the distribution sharpness.

Based on the logit vector $a_n$, we construct a dense routing distribution $p_n$ by applying a softmax function. Concurrently, we compute a Top-$K$ sparse routing distribution $q_n$ by retaining only the top-$K$ elements in $a_n$, setting the remainder to zero, and re-normalizing the retained values. We then blend these two distributions using a balancing coefficient $\lambda$ to obtain the final routing weight vector $\pi_n$:
\begin{subequations}\label{eq:ddc_route}
\begin{align}
p_n &= \mathrm{Softmax}(a_n), \label{eq:ddc_route1}\\
q_n &= \mathrm{TopKSoftmax}(a_n; K), \label{eq:ddc_route2}\\
\pi_n &= \lambda q_n + (1-\lambda)p_n, \label{eq:ddc_route3}
\end{align}
\end{subequations}
\noindent where $p_n, q_n, \pi_n \in \mathbb{R}^{M}$ represent probability vectors that each sum to 1, $K$ is the number of activated experts in the sparse routing path, and $\lambda \in [0,1]$ explicitly controls the sparsity of the final routing weights.

\subsubsection{Position-dependent dynamic convolution}
With routing weights $\pi_{n,m}$, the equivalent convolution parameters at location $n$ are
\begin{equation}
\hat{W}_n=\sum_{m=1}^{M}\pi_{n,m}W_m,\qquad
\hat{b}_n=\sum_{m=1}^{M}\pi_{n,m}b_m,
\label{eq:ddc_equiv}
\end{equation}
\noindent where $\hat{W}_n$ and $\hat{b}_n$ denote the aggregated location-specific convolution kernel and bias, respectively, while $W_m$ and $b_m$ represent the base parameters of the $m$-th expert. Computationally, \eqref{eq:ddc_equiv} is a weighted linear combination of expert parameters, ensuring that the entire process is fully differentiable.

The output at location $n$ and channel $c$ follows standard convolution with location-dependent parameters:
\begin{equation}
y_{n,c} = \sum_{c'=1}^{C} \sum_{\Delta\in\Omega} \hat{W}_{n,c,c',\Delta}\,F_{c',\,n+\Delta} + \hat{b}_{n,c},
\label{eq:ddc_conv}
\end{equation}
where $\Omega$ denotes the set of spatial offsets in the convolution window and $\Delta$ is an offset index; $n+\Delta$ indicates the neighboring position of $n$. $F_{c',\,n+\Delta}$ is the sampled input feature at an offset $\Delta$ around $n$, and $\hat{W}_{n,c,c',\Delta}$ denotes the location-dependent kernel weight.

\subsubsection{Residual encapsulation and post-processing}
Finally, the dynamic convolution operator is integrated into a residual block to ensure stable feature propagation. We sequentially apply Layer Normalization and a GELU activation function to the dynamic convolution output, followed by a residual connection with the original input $F_{in}$:
\begin{equation}
{F}_{out} = F_{in} + \mathrm{GELU}\Big(\mathrm{LN}\big(\mathrm{DDC}(F_{in}, D)\big)\Big),
\label{eq:ddc_block}
\end{equation}
\noindent where $F_{out}$ denotes the final enhanced feature map of the block, $\mathrm{DDC}(\cdot, \cdot)$ is the proposed prior-driven dynamic convolution operator, $\mathrm{LN}(\cdot)$ represents Layer Normalization, and $\mathrm{GELU}(\cdot)$ stands for the Gaussian Error Linear Unit activation function.

\subsection{Shuffle channel fusion block}
At the same scale, IR and VIS features often encode complementary information in different channel subspaces: IR features emphasize target saliency and structural contours, while VIS features contain richer textures and background semantics. If we solely rely on global channel mixing via concatenation followed by a standard $1\times1$ convolution, the cross-modal complementarity might be excessively homogenized. To address this issue, we introduce a channel-wise mixing convolution that performs sliding aggregation along the channel dimension. Coupled with a channel shuffling operation that interleaves the channels of both modalities, this design facilitates fine-grained cross-modal interactions within localized channel neighborhoods, ultimately leading to a more effective fusion.

\begin{figure}[!b]
  \centering
  \includegraphics[width=\linewidth]{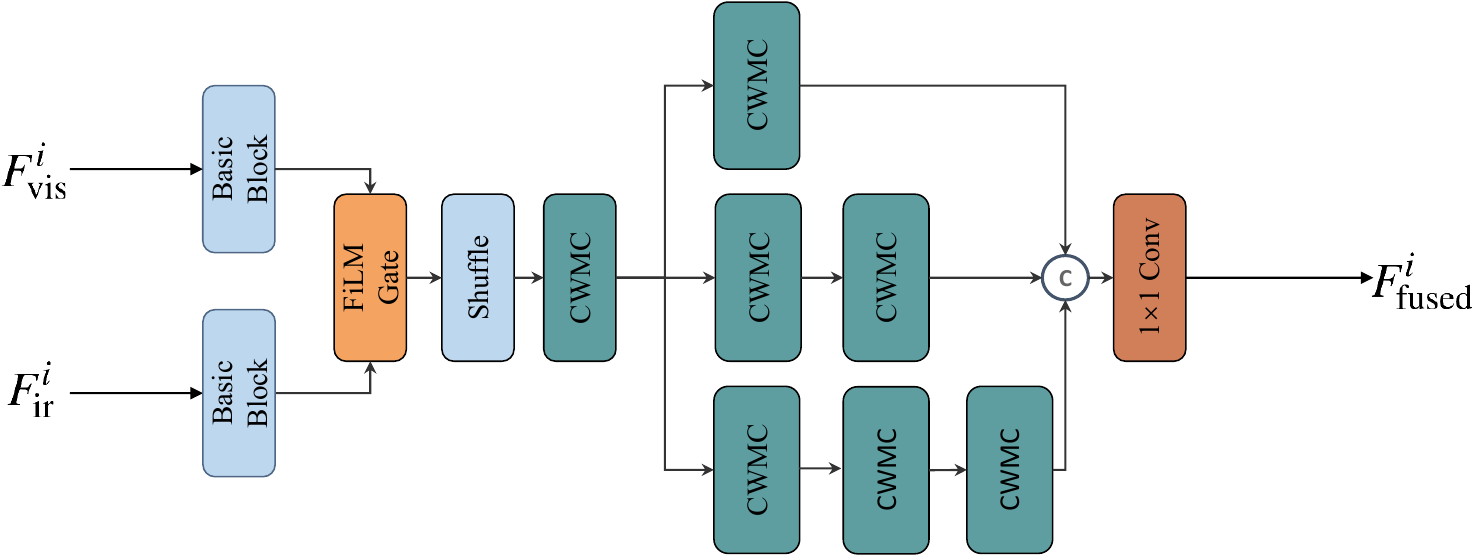}
  \caption{Shuffle Channel Fusion Block (SCFB). FiLM-based cross-modal gating, channel shuffling, local channel mixing, and multi-branch interaction are followed by a $1\times1$ projection.}
  \label{fig:scfb}
\end{figure}

As illustrated in Fig.~\ref{fig:scfb}, the proposed SCFB sequentially performs preprocessing, cross-modal FiLM gating~\cite{Tang_2023}, channel shuffling, and channel-wise mixing convolution (CWMC) with multi-branch interactions, followed by a final $1\times1$ projection to produce the scale-wise fused feature.

\subsubsection{FiLM-based gating modulation}
At the $i$-th scale, we first apply pre-processing and gated modulation to the input features:
\begin{subequations}\label{eq:scfb_pre}
\begin{align}
\bar{F}_{vis}^{i} &= \mathcal{R}\big(F_{vis}^{i}\big), \label{eq:scfb_pre1}\\
\bar{F}_{ir}^{i} &= \mathcal{R}\big(F_{ir}^{i}\big), \label{eq:scfb_pre2}\\
(\tilde{F}_{ir}^{i},\tilde{F}_{vis}^{i}) &= \mathcal{G}\big(\bar{F}_{ir}^{i},\bar{F}_{vis}^{i}\big), \label{eq:scfb_gate}
\end{align}
\end{subequations}
\noindent where $F_{vis}^{i}, F_{ir}^{i} \in \mathbb{R}^{B \times C_i \times H_i \times W_i}$ denote the input feature maps, with $B$, $C_i$, $H_i$, and $W_i$ representing the batch size, channel number, and spatial dimensions at scale $i$, respectively. $\mathcal{R}(\cdot)$ denotes a Basic Block employed for local enhancement, and $\mathcal{G}(\cdot,\cdot)$ represents the gating modulation module.

We implement $\mathcal{G}(\cdot,\cdot)$ via a FiLM-style channel-wise affine modulation. Specifically, we compute a joint global descriptor to predict the scale and bias coefficients for each modality, which are subsequently used to modulate both feature streams:
\begin{subequations}\label{eq:scfb_film}
\begin{align}
z^{i} &= \mathrm{GAP}\big(\mathrm{Cat}[\bar{F}_{ir}^{i},\bar{F}_{vis}^{i}]\big), \label{eq:scfb_film1}\\
\boldsymbol{\phi}^{i} &= \mathrm{MLP}(z^{i}) = [\gamma_{ir}^{i},\beta_{ir}^{i},\gamma_{vis}^{i},\beta_{vis}^{i}], \label{eq:scfb_film2}\\
\tilde{F}_{m}^{i} &= \gamma_{m}^{i}\odot \bar{F}_{m}^{i}+\beta_{m}^{i},\quad m\in\{ir,vis\}, \label{eq:scfb_film3}
\end{align}
\end{subequations}
\noindent where $\mathrm{GAP}(\cdot)$ denotes global average pooling, $\mathrm{Cat}[\cdot,\cdot]$ denotes concatenation along the channel dimension, and $z^{i}$ is the joint global descriptor. The function $\mathrm{MLP}(\cdot)$ represents a multi-layer perceptron. The terms $\gamma_{m}^{i}$ and $\beta_{m}^{i}$ denote the scale and bias coefficients for modality $m$, respectively, which are broadcast along the spatial dimensions, and $\odot$ denotes element-wise multiplication along the channel dimension.

\subsubsection{Channel-wise mixing convolution}
\begin{figure}[htb]
  \centering
  \includegraphics[width=\linewidth]{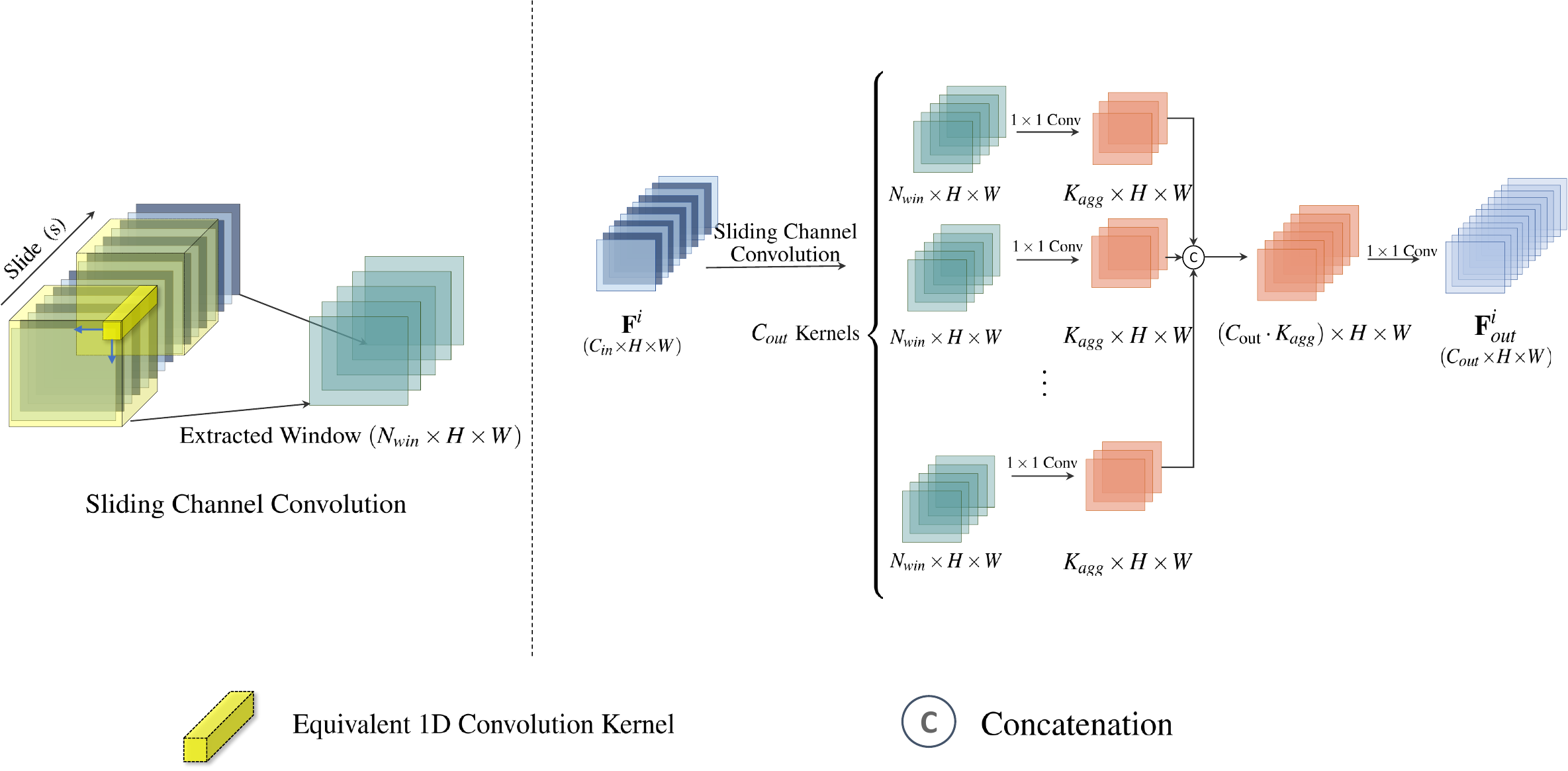}
  \caption{Channel-wise mixing convolution (CWMC). A sliding-window 1D convolution along channels is folded and projected by $1\times1$ layers to obtain local channel interactions.}
  \label{fig:cwmc}
\end{figure}

To enhance cross-modal information interaction, we design a channel-wise mixing convolution (CWMC), denoted as $\mathcal{C}_{k,s}^{i}(\cdot)$. As shown in Fig.~\ref{fig:cwmc}, the key idea is to treat the channel vector at each spatial location $(h,w)$ as a 1D sequence, and perform local sliding-window convolution along the channel dimension with window length $k$ and stride $s$. This local channel aggregation avoids excessive homogenization caused by one-shot full-channel mixing.

Given an input feature $F^{i}\in\mathbb{R}^{B\times C_{\mathrm{in}}\times H\times W}$, we first apply padding along the channel dimension denoted as $p$, and then perform a 1D convolution along the channel dimension for each $(h,w)$. We denote this operation as $\Psi_{k,s}^{\mathrm{ch}}(\cdot)$:
\begin{equation}
\begin{aligned}
Y^{i} &= \Psi_{k,s}^{\mathrm{ch}}\!\left(F^{i}\right)\in\mathbb{R}^{B\times C_{\mathrm{out}}\times N_{\mathrm{win}}\times H\times W},\\
N_{\mathrm{win}} &= \Big\lfloor (C_{\mathrm{in}}+2p-k)/s \Big\rfloor+1,
\end{aligned}
\label{eq:cmix_slide}
\end{equation}
where $\Psi_{k,s}^{\mathrm{ch}}(\cdot)$ denotes sliding-window 1D convolution along the channel dimension shared across spatial locations, $C_{\mathrm{out}}$ is the number of channel kernel groups, and $N_{\mathrm{win}}$ is the response length produced by the sliding windows.

We then fold the window dimension $N_{\mathrm{win}}$ into the channel dimension and employ two standard $1\times1$ convolutions for feature compression and projection:
\begin{subequations}\label{eq:cmix_proj}
\begin{align}
Z^{i} &= \mathrm{Conv}_{1\times1}\!\left(\mathcal{R}(Y^{i})\right), \label{eq:cmix_proj1}\\
F_{\mathrm{out}}^{i} &= \mathrm{Conv}_{1\times1}\!\left(Z^{i}\right). \label{eq:cmix_proj2}
\end{align}
\end{subequations}
where $\mathcal{R}(\cdot)$ denotes the reshape operation that collapses the window dimension into the channel manifold, and $Z^i \in \mathbb{R}^{B \times (C_{\mathrm{out}} \cdot K_{\mathrm{agg}}) \times H \times W}$ represents the intermediate representation. $\mathrm{Conv}_{1\times1}(\cdot)$ refers to the standard pointwise convolution. $F_{\mathrm{out}}^i$ is the final output feature of the proposed operator.

Overall, we denote the operator as
\begin{equation}
F_{\mathrm{out}}^{i}=\mathcal{C}_{k,s}^{i}\!\left(F^{i}\right).
\label{eq:cmix_def}
\end{equation}

\subsubsection{Scale-wise fusion}
As shown in Fig.~\ref{fig:scfb}, we concatenate and shuffle the gated features to interleave modality channels, and then apply channel mixing to obtain an embedding:
\begin{equation}
E^{i}=\mathcal{C}_{k,s}^{(i)}\Big(\mathrm{Shuffle}(\tilde{F}_{vis}^{i},\tilde{F}_{ir}^{i})\Big).
\label{eq:scfb_embed}
\end{equation}
where $\mathrm{Shuffle}(\cdot,\cdot)$ denotes channel shuffling after concatenation to interleave the channels of the two modalities. $\mathcal{C}_{k,s}^{(i)}(\cdot)$ instantiates the channel-wise mixing convolution in \eqref{eq:cmix_def}, and $E^{i}\in\mathbb{R}^{B\times 2C_i\times H_i\times W_i}$ is the interleaved embedding feature.

Given the interleaved embedding $E^{i}$, we employ three parallel channel-mixing branches with varying depths and subsequently project the aggregated features back to $C_i$ channels:
\begin{subequations}\label{eq:scfb_fuse}
\begin{align}
F_1^{i} &= \mathcal{T}_1\big(E^{i}\big),\quad
F_2^{i} = \mathcal{T}_2\big(E^{i}\big),\quad
F_3^{i} = \mathcal{T}_3\big(E^{i}\big), \label{eq:scfb_branches}\\
F_{\mathrm{fused}}^{i} &= \mathrm{Conv}_{1\times1}\Big(\mathrm{Cat}[F_1^{i},F_2^{i},F_3^{i}]\Big).
\label{eq:scfb_fused}
\end{align}
\end{subequations}
\noindent where $\mathcal{T}_1(\cdot)$, $\mathcal{T}_2(\cdot)$, and $\mathcal{T}_3(\cdot)$ denote the respective channel-mixing branches designed to capture multi-scale interaction strengths.

\subsection{Loss functions}
In the absence of ground-truth supervision, we optimize the fusion network using complementary intensity- and structure-preserving objectives. This enables the fused image to retain both prominent thermal targets and rich visible textures. Let $I, V \in \mathbb{R}^{H \times W}$ denote the input IR and VIS images, respectively, and let $F \in \mathbb{R}^{H \times W}$ be the fused output, where $H$ and $W$ represent the spatial dimensions. The overall loss function is formulated as follows:
\begin{equation}
\mathcal{L}_{\mathrm{total}}=\lambda_{1}\mathcal{L}_{\mathrm{int}}+\lambda_{2}\mathcal{L}_{\mathrm{grad}}+\lambda_{3}\mathcal{L}_{\mathrm{ssim}}+\lambda_{4}\mathcal{L}_{\mathrm{struct}},
\label{eq:loss_total}
\end{equation}
\noindent where $\lambda_{1}$ through $\lambda_{4}$ are trade-off coefficients controlling the relative importance of each term.

To encourage the fused image to retain salient intensity responses from both source modalities and prevent prominent targets or significant luminance features from being degraded, we define the intensity preservation term as:
\begin{equation}
\mathcal{L}_{\mathrm{int}}=\frac{1}{HW}\left\|F-\max(I,V)\right\|_{1},
\label{eq:loss_int}
\end{equation}
\noindent where $\|\cdot\|_{1}$ denotes the $\ell_{1}$ norm, and $\max(\cdot)$ indicates the pixel-wise maximum selection between $I$ and $V$.

To enhance edges and fine textures, we compel the gradients of the fused image to align with the maximum gradient responses from the two source images:
\begin{equation}
\mathcal{L}_{\mathrm{grad}}=\frac{1}{HW}\left\|\nabla F-\max(\nabla I,\nabla V)\right\|_{1},
\label{eq:loss_grad}
\end{equation}
\noindent where $\nabla(\cdot)$ denotes the gradient magnitude computed via the Sobel operator, and $\max(\cdot)$ is defined as in \eqref{eq:loss_int}.

We further enforce structural consistency between the fused image and both source modalities using weights adaptively derived from their global gradient strengths:
\begin{subequations}\label{eq:loss_ssim_all}
\begin{align}
\mathcal{L}_{\mathrm{ssim}}&=1-\Big(\omega_I\,\mathrm{SSIM}(I,F)+\omega_V\,\mathrm{SSIM}(V,F)\Big), \label{eq:loss_ssim}\\
\omega_I&=\frac{\mu(\nabla I)}{\mu(\nabla I)+\mu(\nabla V)},\qquad\omega_V=1-\omega_I, \label{eq:loss_ssim_w}
\end{align}
\end{subequations}
\noindent where $\mathrm{SSIM}(\cdot,\cdot)$ denotes the structural similarity index~\cite{Wang_2004_SSIM}, and $\mu(\cdot)$ computes the global mean of a given gradient map.

To explicitly encourage complementary structure injection, we define two residual maps:
\begin{equation}
R_V=\left|F-V\right|,\qquad R_I=\left|F-I\right|,
\label{eq:loss_residuals}
\end{equation}
\noindent where $R_V$ and $R_I$ measure the absolute residuals of the fused image relative to the VIS and IR images, respectively. Subsequently, we constrain the gradients of the residual structures relative to one modality to match the gradients of the other modality:
\begin{equation}
\mathcal{L}_{\mathrm{struct}}=\frac{1}{HW}\Big(\left\|\nabla R_V-\nabla I\right\|_1+\left\|\nabla R_I-\nabla V\right\|_1\Big),
\label{eq:loss_struct}
\end{equation}
\noindent where $\nabla(\cdot)$ represents the aforementioned gradient magnitude operator. This term mitigates structural inconsistencies and artifacts, thereby promoting a more thorough cross-modal structure injection into $F$.
\section{Experiments}
\label{sec:experiments}

\subsection{Experimental settings}
To evaluate the proposed approach, we utilize three widely adopted datasets in the field of infrared and visible image fusion: MSRS~\cite{Tang_2022_PIAFusion}, M3FD~\cite{Liu_2022}, and TNO~\cite{Toet_2012_ProgressColorNightVision}. Specifically, the MSRS dataset provides 1444 aligned infrared and visible image pairs. From this, a subset of 1083 pairs serves as our training corpus, from which $128 \times 128$ spatial patches are randomly extracted during preprocessing. The held-out 361 pairs constitute the primary test dataset. Furthermore, to rigorously examine the cross-domain adaptability and generalization capabilities of our network, we conduct zero-shot evaluations on the external M3FD and TNO datasets.

The entire framework is developed using the PyTorch library, with all computational procedures executed on an infrastructure featuring one NVIDIA RTX 4090 GPU. Optimization is driven by the Adam algorithm over a span of 60 epochs, utilizing a batch size of 32. We initialize the learning rate at 1.7e-4, subsequently modulating it via a cosine annealing strategy preceded by a linear warm-up phase. To achieve an optimal equilibrium among the constituent objective functions, the weighting coefficients $\lambda_{1}$, $\lambda_{2}$, $\lambda_{3}$, and $\lambda_{4}$ are empirically assigned values of 4.0, 24.0, 0.5, and 4.5, respectively.

To benchmark our network's performance, we compare it against eight recent advanced VIF architectures: EMMA~\cite{Zhao_2024}, SwinFusion~\cite{Ma_2022}, T2EA~\cite{Huang_2025_T2EA}, WaveFusion~\cite{Wang_2025_WaveFusion}, CDDFuse~\cite{Zhao_2023_CDDFuse}, MaeFuse~\cite{Li_2025_maefuse}, SPDFusion~\cite{Xiao_2025}, and TDFusion~\cite{Bai_2025}. Quantitative assessment is carried out using a suite of five established evaluation criteria: Entropy (EN), Spatial Frequency (SF), Average Gradient (AG), Standard Deviation (SD), and Visual Information Fidelity (VIF). Across these chosen indices, a larger numerical score directly correlates with enhanced fusion quality. For all metric tables, the bold and underlined values represent the optimal and second-best results, respectively.

\subsection{Quantitative comparison}

As summarized in Table~\ref{tab:m3fd_msrs_tno}, \MethodName demonstrates consistent superiority across three datasets. First, it achieves the best performance on EN, SF, and AG. By maximizing these indicators of information capacity and detail activity, the proposed strategy robustly aggregates complementary multimodal features and enhances texture responses. Second, regarding SD, a metric reflecting global contrast, \MethodName ranks first on MSRS and M3FD and second on TNO. This proves its effectiveness in highlighting salient IR targets while preserving VIS background visibility and gray-level separability. Third, for the perceptual metric VIF, our method ranks second on MSRS, confirming high in-domain visual fidelity. The performance gap on M3FD and TNO relative to the top method is primarily attributed to cross-dataset domain shifts.

\begin{table}[t]
  \centering
  \normalsize
  \caption{Quantitative evaluation on M3FD, MSRS, and TNO datasets.}
  \label{tab:m3fd_msrs_tno}
  \resizebox{\textwidth}{!}{%
  \begin{tabular}{lccccc ccccc ccccc}
    \toprule
    Method
      & \multicolumn{5}{c}{M3FD}
      & \multicolumn{5}{c}{MSRS}
      & \multicolumn{5}{c}{TNO} \\
    \cmidrule(lr){2-6} \cmidrule(lr){7-11} \cmidrule(lr){12-16}
      & EN & SF & AG & SD & VIF
      & EN & SF & AG & SD & VIF
      & EN & SF & AG & SD & VIF \\
    \midrule
    EMMA       & 6.924 & 15.227 & \underline{5.341} & 38.255 & 0.769  & 6.723 & 11.559 & 3.788 & \underline{44.590} & 0.974  & 7.240 & 12.650 & 5.165 & \textbf{48.467} & 0.692 \\
    SwinFusion & 6.790 & 13.685 & 4.615 & 35.834 & \underline{0.774} & 6.621 & 11.089 & 3.566 & 42.998 & 0.990  & 6.898 & 11.679 & 4.511 & 40.864 & 0.724 \\
    T2EA       & 6.643 & 10.452 & 3.894 & 29.664 & 0.609  & 6.316 & 8.477  & 2.909 & 34.642 & 0.779  & 7.039 & 9.874  & 4.125 & 39.728 & 0.636 \\
    WaveFusion & 6.983 & \underline{15.889} & 5.226 & 37.224 & 0.772  & 6.720 & \underline{11.999} & \underline{3.988} & 42.936 & \underline{1.019} & 7.224 & \underline{14.534} & 5.356 & 45.707 & \underline{0.748} \\
    CDDFuse    & 6.899 & 14.775 & 4.874 & 37.236 & \textbf{0.793} & 6.701 & 11.557 & 3.748 & 43.382 & \textbf{1.051} & 7.151 & 13.959 & 5.078 & 46.581 & \textbf{0.752} \\
    MaeFuse    & 6.966 & 8.831  & 3.639 & 35.410 & 0.477  & 6.605 & 9.407  & 3.432 & 38.383 & 0.711  & 6.967 & 8.762  & 3.981 & 37.570 & 0.495 \\
    SPDFusion  & 6.872 & 14.952 & 5.144 & 35.861 & 0.748  & 6.682 & 11.460 & 3.800 & 42.683 & 0.988  & 7.145 & 14.075 & \underline{5.541} & 44.362 & 0.679 \\
    TDFusion   & \underline{7.063} & 14.365 & 4.947 & \underline{39.736} & 0.769  & \underline{6.738} & 11.305 & 3.736 & 42.950 & 1.001  & \underline{7.248} & 13.313 & 5.089 & 46.888 & 0.699 \\
    Ours       & \textbf{7.179} & \textbf{18.007} & \textbf{6.238} & \textbf{43.908} & 0.764  & \textbf{6.831} & \textbf{13.199} & \textbf{4.398} & \textbf{47.544} & \underline{1.026} & \textbf{7.314} & \textbf{15.176} & \textbf{6.604} & \underline{48.136} & 0.658 \\
    \bottomrule
  \end{tabular}%
  }
\end{table}

\begin{table}[!tbp]
  \centering
  \caption{Quantitative results on semantic segmentation.}
  \label{tab:seg-results}
  \resizebox{\textwidth}{!}{%
  \begin{tabular}{lcccccccccccc}
    \toprule
    Model & Unl & Car & Per & Bik & Cur & CS & GD & CC & Bu & mIoU & aAcc & mAcc \\
    \midrule
    EMMA       & 97.90 & 86.17 & 66.66 & 67.25 & 46.06 & 52.00 & \underline{43.94} & 53.89 & 68.70 & 64.73 & 97.87 & 72.55 \\
    SwinFusion & 97.91 & \underline{86.26} & 67.77 & 67.50 & 44.22 & 51.80 & 41.66 & 53.55 & 69.38 & 64.45 & 97.88 & 72.08 \\
    T2EA       & 97.91 & 85.84 & 68.11 & \textbf{68.52} & 44.82 & 51.41 & 41.45 & 53.17 & \textbf{71.47} & 64.74 & 97.88 & 72.47 \\
    WaveFusion & \underline{97.94} & \textbf{86.59} & 68.52 & 67.76 & \underline{46.68} & 53.36 & 41.47 & 52.21 & 67.51 & 64.67 & \underline{97.91} & 72.77 \\
    CDDFuse    & 97.91 & 86.11 & 67.44 & 66.78 & 45.62 & 52.06 & 43.85 & \underline{54.01} & \underline{71.14} & 64.99 & 97.87 & 73.02 \\
    MaeFuse    & 97.89 & 86.13 & \textbf{68.63} & 67.88 & 44.95 & 48.92 & 39.72 & 52.66 & 68.36 & 63.90 & 97.86 & 71.82 \\
    SPDFusion  & 97.92 & 85.91 & 67.89 & 68.11 & 44.61 & \underline{54.46} & 39.48 & 53.27 & 69.96 & 64.62 & 97.89 & 72.65 \\
    TDFusion   & 97.93 & 86.25 & \textbf{68.63} & 67.46 & 44.34 & 53.92 & 42.32 & 53.20 & 71.11 & \underline{65.02} & 97.90 & \underline{73.12} \\
    Ours       & \textbf{97.99} & 86.15 & \underline{68.57} & \underline{68.43} & \textbf{48.71} & \textbf{56.01} & \textbf{48.98} & \textbf{55.79} & 67.63 & \textbf{66.47} & \textbf{97.96} & \textbf{74.43} \\
    \bottomrule
  \end{tabular}%
  }
\end{table}

\subsection{Qualitative comparison}
\begin{figure}[!htbp]
  \centering
  \includegraphics[width=\textwidth,height=0.39\textheight,keepaspectratio]{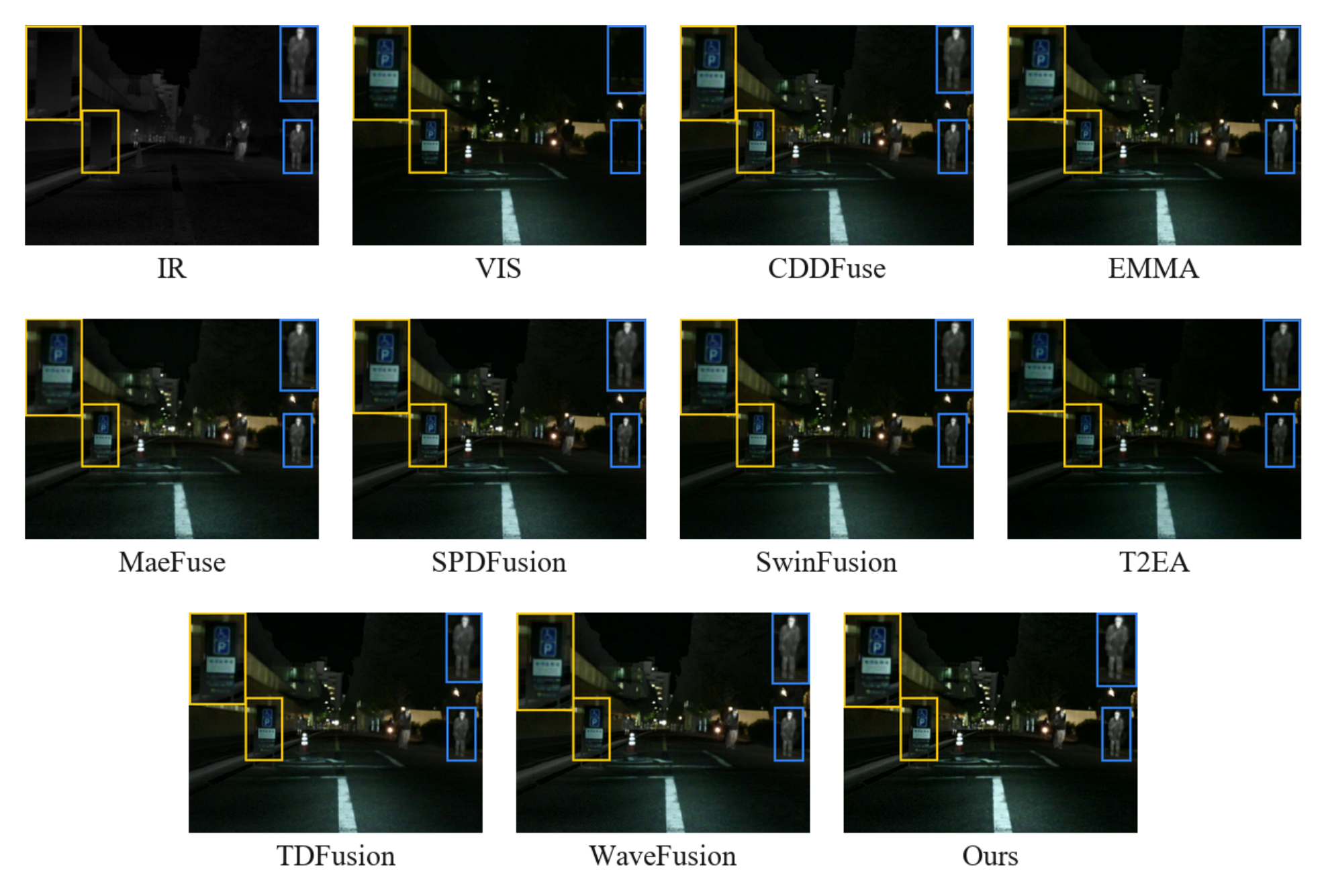}\par\medskip
  \includegraphics[width=\textwidth,height=0.39\textheight,keepaspectratio]{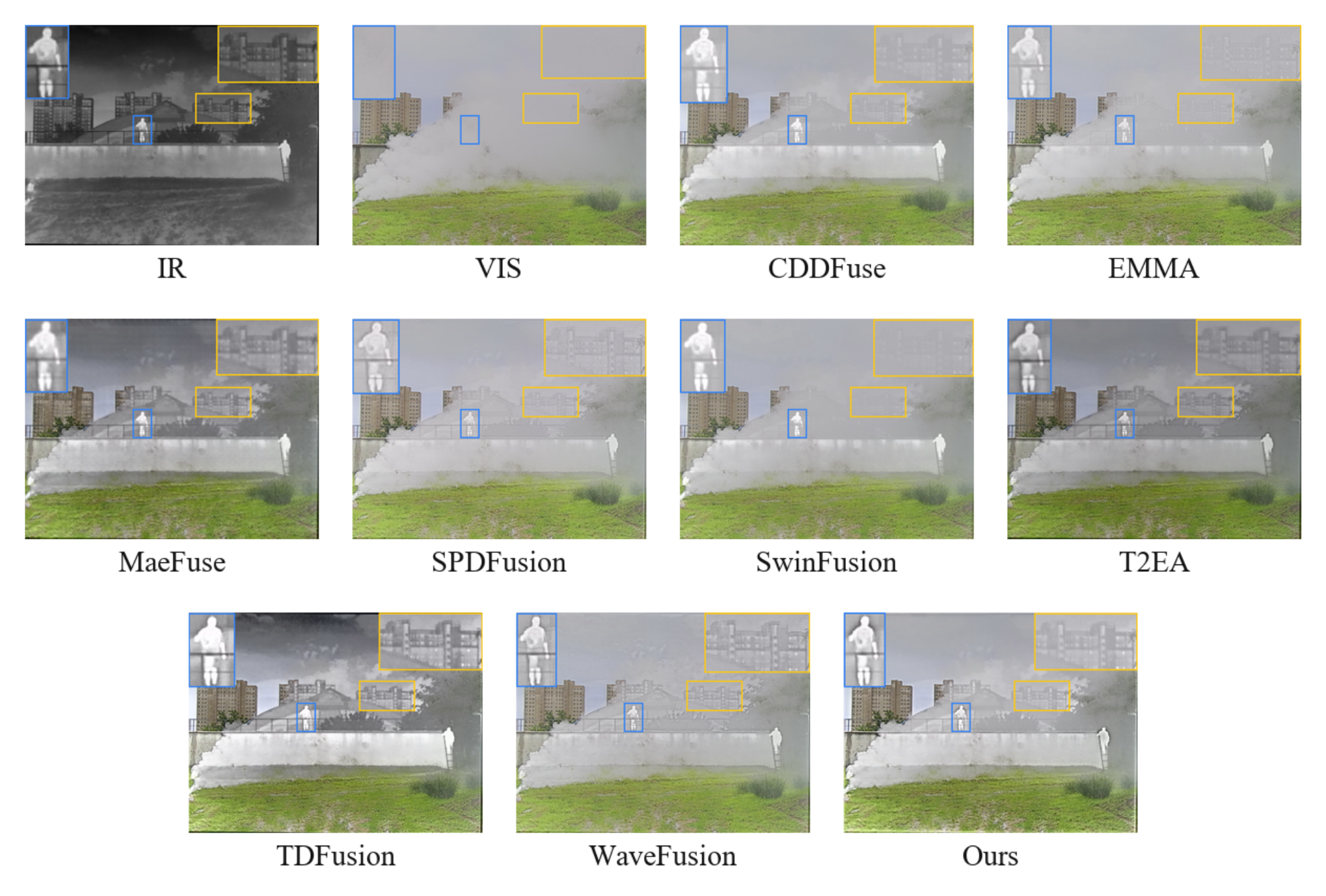}\par
  \caption{Qualitative fusion comparison on MSRS and M3FD. Columns show VIS, IR, and fused results of CDDFuse, EMMA, MaeFuse, SPDFusion, SwinFusion, T2EA, TDFusion, WaveFusion, and Ours; boxed regions are enlarged.}
  \label{fig:qualitative}
\end{figure}

As illustrated in Fig.~\ref{fig:qualitative}, the visual comparison on the MSRS and M3FD datasets reveals distinct limitations in existing state-of-the-art methods. Specifically, CDDFuse, EMMA, and SwinFusion exhibit insufficient overall contrast and fail to adequately render structures that are occluded or situated in low-light regions, often resulting in diminished visual contrast and compromised situational awareness. Meanwhile, MaeFuse and WaveFusion are prone to introducing noticeable visual artifacts and noise, particularly at the boundaries of thermal targets, which disrupt the coherence of the fused image. Furthermore, T2EA suffers from insufficient detail preservation, leading to blurred textures in complex background areas. In contrast, our proposed \MethodName effectively overcomes these drawbacks by achieving superior global contrast and structure recovery; it sharply delineates occluded details and salient thermal targets without introducing visual artifacts, yielding the most visually informative and natural fusion results among all compared methods.

\subsection{Downstream applications}
To verify the practical value of fused images for downstream tasks, we conduct semantic segmentation experiments on the MSRS dataset, which includes nine categories: background, speed bump, cone, guardrail, curve, bicycle, person, car-stop, and car. The experiment consists of two stages. First, we generate fused results using our method and mainstream fusion baselines. Second, we train a DeepLabV3+~\cite{Chen_2018_DeepLabv3Plus} network on the MMSegmentation~\cite{mmseg2020} platform with these fused images as input. We initialize the backbone with ImageNet-pretrained weights to improve convergence speed and generalization stability. Specifically, we use the SGD optimizer with an initial learning rate of $7\times10^{-3}$, a momentum of 0.9, and a weight decay of $5\times10^{-4}$. We adopt the PolyLR schedule to decay the learning rate to $1\times10^{-4}$ over 2750 iterations. We use 1083 image pairs for training and 361 image pairs for testing. Model performance is evaluated by mean IoU (mIoU), mean class accuracy (mAcc), and overall pixel accuracy (aAcc).

\begin{figure}[!htbp]
  \centering
  \includegraphics[width=\textwidth]{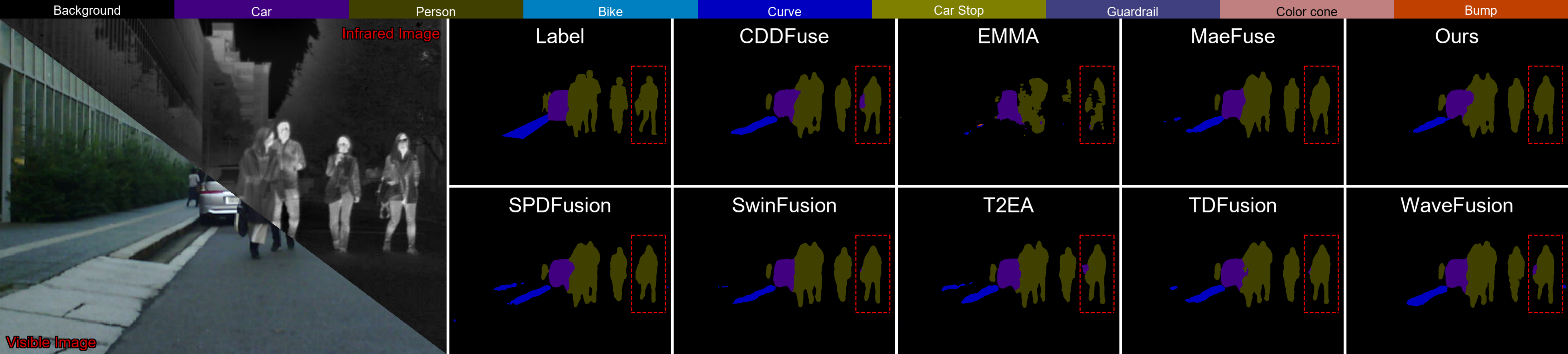}\par\medskip
  \includegraphics[width=\textwidth]{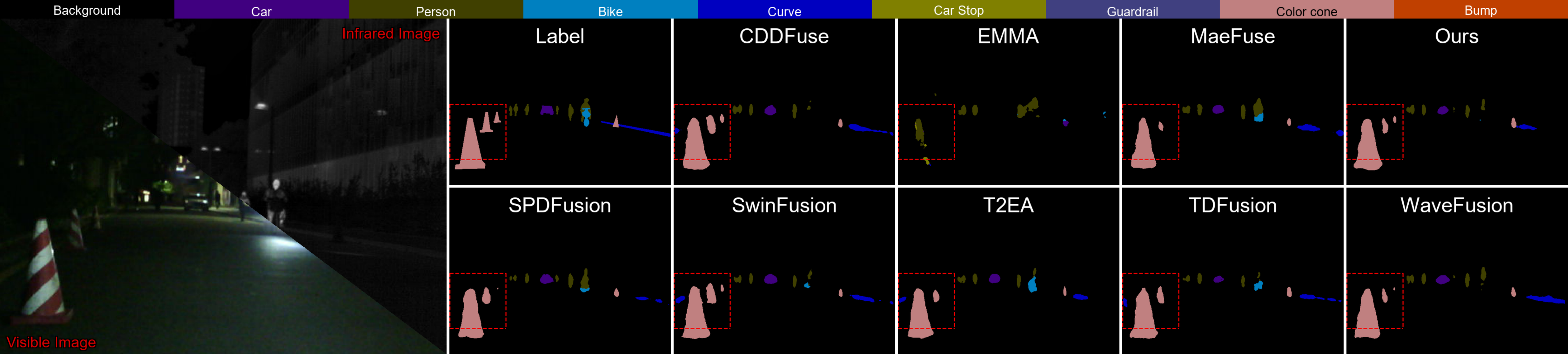}
  \caption{Qualitative segmentation comparison in daytime (top) and nighttime (bottom) scenes. Columns show source VIS/IR images and segmentation maps from different fusion methods; boxed regions highlight differences. VIS: visible; IR: infrared.}
  \label{fig:seg_qualitative}
\end{figure}

As reported in Table~\ref{tab:seg-results}, \MethodName achieves the best overall segmentation performance, securing the highest mIoU, aAcc, and mAcc among all competitors. At the category level, our approach leads across most classes, performing exceptionally well on safety-critical categories for autonomous driving, such as Bike and Person. These improvements demonstrate that the proposed self-evolving auxiliary priors successfully preserve discriminative target structures and fine-grained textures, thereby providing highly informative representations for downstream segmentation.

Fig.~\ref{fig:seg_qualitative} illustrates qualitative segmentation comparisons across representative daytime and nighttime scenarios. Focusing on the encircled pedestrian examples in the daytime scene (top), our method successfully avoids erroneous classification regions, thereby producing more complete and accurate segmentation masks compared to several competing methods that exhibit fragmented or missing regions, especially around object boundaries. In the nighttime scene (bottom), the advantage of \MethodName becomes more pronounced. As visually depicted, our model demonstrates superior preservation of the contours of the three road cones. This is attributed to the fact that the fused images generated by our method retain clearer target contours and better contrast under low-illumination conditions, enabling the segmentation network to correctly delineate objects that are partially or entirely missed by other approaches. These visual results are consistent with the quantitative findings in Table~\ref{tab:seg-results}, confirming that the high-quality fusion produced by \MethodName effectively benefits downstream semantic understanding tasks.

\FloatBarrier

\subsection{Ablation study}
To further validate the proposed design, we conduct ablation experiments on the MSRS dataset. Unless otherwise specified, we follow the full model and only modify the targeted component.

\subsubsection{Basic module ablation}
Table~\ref{tab:ablation_msrs_part1} investigates the contributions of the DDCB and the SCFB. In Setting~I, both modules are removed, and the model reverts to using the Restormer-CNN block~\cite{Zhao_2024} as the baseline operator, discarding any prior-conditioned dynamic kernels. To isolate the effect of each module, the DDCB and the SCFB are individually replaced by the Restormer-CNN block in Setting~II and Setting~III, respectively. In Setting~IV, the DDCB is similarly replaced by the Restormer-CNN block, but the APG priors are explicitly injected via channel concatenation. Setting~V denotes the full model.

As shown in Table~\ref{tab:ablation_msrs_part1}, the full model (Setting~V) achieves the best overall performance across all metrics. The most pronounced degradation appears in Setting~I, indicating that the DDCB and the SCFB provide complementary gains and are most effective when jointly utilized. Removing either the DDCB in Setting~II or the SCFB in Setting~III also leads to consistent performance drops, confirming the importance of both modules. Moreover, the concatenation-based prior injection in Setting~IV underperforms the DDCB-based design, suggesting that conditioning local operators on the APG priors is more effective than naive concatenation.

\subsubsection{Ablation on adaptive auxiliary prior usage}
Table~\ref{tab:ablation_msrs_part2} examines the update rule of the auxiliary prior in the APG module. In Setting~I, cross-scale inheritance is removed and the dictionary is updated only from current-level proposals. In Setting~II, the proposal update is removed and only inherited priors are retained. Settings~III and IV modify the dictionary initialization by sharing the learnable initial prior $D_0$ between the IR and VIS branches or freezing $D_0$ during training. Setting~V corresponds to the full model.

\begin{table}[b]
  \centering
  \normalsize
  {\renewcommand{\arraystretch}{1.12}%
  \setlength{\tabcolsep}{4pt}%
  \caption{Ablation study on MSRS (Basic Module Ablation).}
  \label{tab:ablation_msrs_part1}
  \begin{tabular}{rlccccc}
    \toprule
    & \textbf{Ablation Setting} & \textbf{EN} & \textbf{SF} & \textbf{AG} & \textbf{SD} & \textbf{VIF} \\
    \midrule
    I    & w/o Fusion \& DynConv     & 6.769 & 12.490 & 4.093 & 44.699 & 0.990 \\
    II   & w/o DynConv          & 6.797 & 12.805 & 4.262 & 44.800 & 1.011 \\
    III  & w/o Fusion          & 6.794 & 12.753 & 4.133 & 46.100 & 1.006 \\
    IV   & Concat auxiliary prior            & 6.827 & 12.362 & 4.163 & 46.373 & 0.993 \\
    V    & Ours (Full)                         & \textbf{6.831} & \textbf{13.199} & \textbf{4.398} & \textbf{47.544} & \textbf{1.026} \\
    \bottomrule
  \end{tabular}
  }%
\end{table}

As shown in Table~\ref{tab:ablation_msrs_part2}, the full model (Setting~V) yields the best overall results. Using a single update source, as in Settings~I and II, consistently degrades performance, indicating that cross-scale inheritance and current-level proposals are jointly required to maintain stability while preserving input adaptivity. In addition, altering the learnable initial prior $D_0$ by sharing or freezing it, as in Settings~III and IV, yields inferior results compared to Setting~V, suggesting that modality-aware and learnable initialization further improves the performance ceiling.

\subsubsection{Ablation of the fusion module}
Table~\ref{tab:ablation_fusion_module} presents the ablation results concerning the design of the SCFB. In Setting~I, the multi-branch interaction is simplified to a single fusion branch. In Setting~II, the overall topology is preserved while the specialized convolution operators are replaced with standard convolutions. Setting~III corresponds to the full model.

\begin{table}[t]
  \centering
  \normalsize
  {\renewcommand{\arraystretch}{1.12}%
  \setlength{\tabcolsep}{4pt}%
  \caption{Ablation study on MSRS (Ablation on adaptive auxiliary prior usage).}
  \label{tab:ablation_msrs_part2}
  \begin{tabular}{rlccccc}
    \toprule
    & \textbf{Ablation Setting} & \textbf{EN} & \textbf{SF} & \textbf{AG} & \textbf{SD} & \textbf{VIF} \\
    \midrule
    I    & Proposal-Only                       & 6.795 & 12.797 & 4.196 & 46.975 & 1.004 \\
    II   & History-Only                        & 6.796 & 12.680 & 4.171 & 46.236 & 0.975 \\
    III  & Shared InitDict                     & 6.817 & 12.849 & 4.166 & 47.465 & 1.016 \\
    IV   & Freeze InitDict                     & 6.808 & 13.149 & 4.361 & 47.379 & 1.019 \\
    V    & Ours (Full)                         & \textbf{6.831} & \textbf{13.199} & \textbf{4.398} & \textbf{47.544} & \textbf{1.026} \\
    \bottomrule
  \end{tabular}
  }%
\end{table}

\begin{table}[t]
  \centering
  \normalsize
  {\renewcommand{\arraystretch}{1.12}%
  \setlength{\tabcolsep}{4pt}%
  \caption{Ablation study on MSRS (Fusion-module ablation).}
  \label{tab:ablation_fusion_module}
  \begin{tabular}{rlccccc}
    \toprule
    & \textbf{Ablation Setting} & \textbf{EN} & \textbf{SF} & \textbf{AG} & \textbf{SD} & \textbf{VIF} \\
    \midrule
    I    & single-branch                       & \textbf{6.843} & 12.598 & 4.162 & 47.226 & 0.977 \\
    II   & Conv module                         & 6.833 & 12.502 & 4.126 & 45.693 & 0.994 \\
    III  & Ours (Full)                        & 6.831 & \textbf{13.199} & \textbf{4.398} & \textbf{47.544} & \textbf{1.026} \\
    \bottomrule
  \end{tabular}
  }%
\end{table}

As shown in Table~\ref{tab:ablation_fusion_module}, Settings~I and II achieve marginally higher EN but lead to a consistent degradation in SF, AG, and VIF relative to Setting~III. Overall, the full SCFB design in Setting~III provides a more balanced fusion, demonstrating that fine-grained cross-modal interaction is essential for producing perceptually informative fused images.

\section{Conclusion}
\label{sec:conclusion}
In this paper, we present \MethodName, which builds self-evolving internal priors to guide fusion without relying on external auxiliary models. Our main contributions include: (1) the Adaptive Prior Generator, which progressively evolves a compact set of priors across scales to provide stable texture-to-semantic guidance; (2) the Prior-driven Dynamic Convolution Block, which converts these priors into instance-adaptive operators for local feature modulation; and (3) the Shuffle Channel Fusion Block, which promotes fine-grained cross-modal interaction via channel shuffling. 

Extensive evaluations on MSRS, M3FD, and TNO datasets demonstrate \MethodName's superiority in both visual quality and objective metrics. Furthermore, improved downstream segmentation performance validates our intrinsic-prior-guided paradigm. Notably, this self-contained approach offers the community a novel shift from relying on external pre-trained models toward internally generated, data-driven prior modeling. However, performance fluctuates across datasets due to inherent distribution shifts. Future work will explore more robust prior generation and interaction mechanisms to mitigate cross-dataset domain gaps and enhance transferability.

\section*{Acknowledgements}
This work was supported by NSFC of China under Grant 62301432 and Grant 62306240, the Natural Science Basic Research Program of Shaanxi under Grant QCYRCXM-2023-057, the Fundamental Research Funds for Central Universities, and Guangdong Basic and Applied Basic Research Foundation under Grant 2025A1515011119.

\section*{CRediT authorship contribution statement}
\textbf{Zhenyu Sun}: Conceptualization, Methodology, Software, Formal analysis, Investigation, Project administration, Writing--original draft.
\textbf{Luobin Zhang}: Validation, Visualization, Data curation.
\textbf{Axi Niu}: Project administration, Writing--review and editing.
\textbf{Haishen Wang}: Formal analysis, Software.
\textbf{Qingsen Yan}: Writing--review and editing, Resources.

\section*{Declaration of generative AI and AI-assisted technologies in the manuscript preparation process}
During the preparation of this work, the authors used ChatGPT to improve the language and readability of parts of the manuscript and to assist with organizing and synthesizing the literature relevant to this study. The tool was used only as an aid and not as a substitute for the authors’ critical assessment. After using this tool, the authors reviewed and edited the content as needed, verified the accuracy of all information , including manually validating all references, and take full responsibility for the content of the published article.

\section*{Declaration of competing interests}
The authors declare that they have no known competing financial interests or personal relationships that could have appeared to influence the work reported in this paper.

\bibliographystyle{elsarticle-num} 
\bibliography{refs}

\end{document}